\pdfoutput=1
\documentclass[notoc]{kling-keye}

\usepackage{multirow}
\usepackage{makecell}
\usepackage{threeparttable}
\usepackage{mathtools}

\title{VibeVideoAgent: Closing the Perceptual Loop\\for Code-Driven Video Generation}
\author{\normalfont\setlength{\tabcolsep}{1.1em}%
  \begin{tabular}{@{}cccc@{}}
    Xuanyu Zheng$^{*}$ & Luoyuan Zhang$^{*}$ & Yancheng Long$^{*}$ & Haoyang Xu$^{*}$
  \end{tabular}\\
  \begin{tabular}{@{}ccccc@{}}
    Muxi Diao & Yuzhe Chen & Kaiyu Jiang & Bin Wen$^{\dagger}$ & Tingting Gao
  \end{tabular}}
\affiliation{KlingAI Research}
\projectlinks{}
\headertext{VibeVideoAgent: Closing the Perceptual Loop for Code-Driven Video Generation}
\reportdate{November 25, 2026}
\reportabstract{Code-driven video generation asks a coding agent to write and run rendering programs. This gives exact control over typography, data visualizations, synchronized audio, and multi-shot scenes, and it leaves an editable project behind. A program that exits cleanly can still render overlapping labels, a camera aimed away from its subject, broken object interactions, or narration that drifts out of sync, and none of these raise a runtime error. Feeding rendered frames and audio back into the coding session bloats its long-horizon context, while critiquing only the finished full-resolution render turns early design mistakes into costly rewrites. We introduce \textbf{VibeVideoAgent}, a training-free framework that lets coding agents inspect their drafts throughout production. Multimodal perception runs in short-lived contexts kept apart from the coder and returns compact, grounded Findings. Seven primitives operate on a unified multi-clock representation of assets and drafts: they explore media from coarse to fine, verify quoted requirements with evidence from every required modality, compare a draft with a reference or an earlier revision on separate clocks, and measure stream signals locally. A lightweight staged protocol has the agent test its riskiest decision on a minimal preview first. On all 239 VibeVideoBench tasks, with matched agent runtimes and a fixed perception model, VibeVideoAgent improves holistic audiovisual quality (L3) for Claude Opus~5.5 ($4.03 \to 4.43$, while L2 rubric compliance rises from $98.87\%$ to $99.35\%$), GPT-6 Astra ($4.03 \to 4.27$), and Claude Sonnet~5.5 ($3.97 \to 4.10$), with the largest gains in audio--visual synchronization (up to $+0.56$). Controlled ablations show that each component contributes, and trajectory audits show that the granularity of code edits governs the trade-off between rubric adherence and perceptual quality.
}
\hypersetup{pdftitle={VibeVideoAgent: Closing the Perceptual Loop for Code-Driven Video Generation},
  pdfauthor={Xuanyu Zheng, Luoyuan Zhang, Yancheng Long, Haoyang Xu, Muxi Diao, Yuzhe Chen, Kaiyu Jiang, Bin Wen, Tingting Gao},
  pdfkeywords={Vibe Video, Video Generation, Video Understanding}}

\begin{document}
\maketitle
\makeatletter\global\@topnum\z@\makeatother
\begingroup
\renewcommand{\thefootnote}{\fnsymbol{footnote}}
\footnotetext[1]{\,Equal contribution.}
\footnotetext[2]{\,Corresponding author: Bin Wen.}
\endgroup
\reportcontents

\section{Introduction}
\label{sec:introduction}

\begin{figure*}[t]
  \centering
  \includegraphics[width=\linewidth]{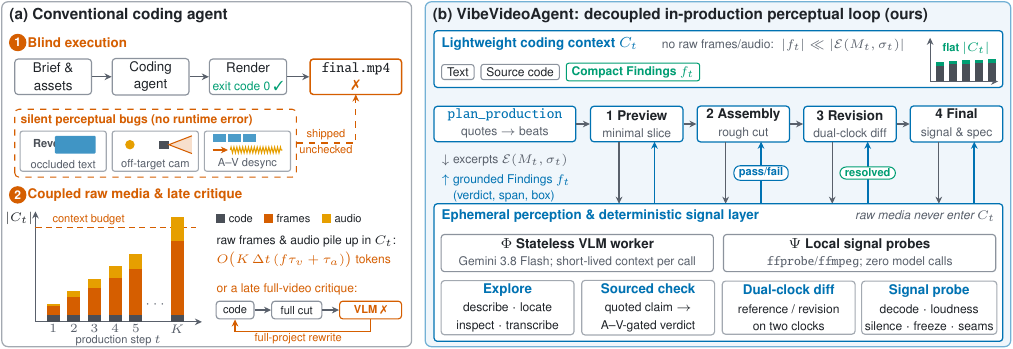}
  \caption{\textbf{Closing the perceptual loop in code-driven video generation.} \textbf{(a)}~A conventional coding agent either ships unchecked renders, so visual, spatial, and sync bugs survive a clean exit code, or puts raw frames and audio into its context $C_t$ and critiques only the final video, which leads to whole-project rewrites. \textbf{(b)}~VibeVideoAgent keeps raw media out of $C_t$: a stateless VLM worker $\Phi$ and local probes $\Psi$ return compact, grounded Findings $f_t$ at every stage, from a minimal preview to final delivery.}
  \label{fig:teaser}
\end{figure*}

Pixel-space video generators~\citep{polyak2024moviegen,kong2024hunyuanvideo,yang2025cogvideox,wan2025wan} and agentic pipelines built on them~\citep{wu2025movieagent} produce compelling textures and motion, but they offer little control when a creator needs exact typography, faithful mathematical or data charts, strict multi-shot continuity, precise audio--visual synchronization, or fine-grained edits afterward. \emph{Code-driven video generation}, often called \emph{Vibe Video}~\citep{diao2026vibevideobench,chen2026vibeprevisbench,chen2026code2video}, treats video creation as programming instead. Given a natural-language brief and raw assets, a coding agent writes and runs rendering software (HTML/Canvas, React/Remotion, Cairo, Manim, FFmpeg audio pipelines, or Blender Python) that yields both a playable video and an editable source project. Because every visual element, camera curve, and audio cue is defined in code, the approach covers both 2D audiovisual motion design~\citep{diao2026vibevideobench,zhang2026animweb} and 4D cinematic previsualization~\citep{chen2026vibeprevisbench,wang2026cinerecon}.

Coding agents write such programs fluently, yet they often fail at authoring video. In software engineering~\citep{yang2024sweagent}, defects usually surface as compiler errors or failing tests. Video authoring has three problems that such feedback does not reach (Figure~\ref{fig:teaser}a). First, \emph{a successful run says little about what the viewer sees}. A script can exit with code zero and write a valid MP4 in which a character holds a backpack in hand instead of wearing it, a 3D camera orbits away from the protagonist, labels collide, or narration lags two seconds behind the chart it explains. Second, \emph{showing raw media to the coder pollutes its context}. Production involves reference footage, design stills, partial shot previews, and several rough cuts, and appending their frames and audio to the agent's history grows the context with every render and degrades long-horizon code maintenance~\citep{liu2024lost,long2026videonexus}. Third, \emph{critique at the end of the pipeline comes too late and cannot follow changes across versions}. When only the finished full-resolution render is reviewed, a basic error in layout, rigging, or timing forces a rewrite of the whole project. Code edits also shift cut timings, so re-inspecting a single clip can accept an unverified fix or miss a regression in a neighboring shot.

We address these problems with \textbf{VibeVideoAgent}, a training-free framework that gives coding agents perceptual feedback while they produce a video (Figure~\ref{fig:teaser}b). The coder never receives raw media. Dense audiovisual observation is delegated to short-lived perception calls, which return compact \emph{Findings} grounded in space and time. This decoupling follows VideoNexus~\citep{long2026videonexus}, which keeps video perception outside a planner's context when answering questions about a fixed video. Authoring adds drafts that change with every edit and live on several clocks, requirements that span audio and vision, and repairs whose effect must be checked across versions. A unified \emph{multi-clock draft representation} addresses read-only reference assets, zero-duration design stills, shot previews placed at an offset in the film, and full cuts in the same way, and binds each observation to hashes of the source code behind it. On top of it, seven primitives serve four roles: (i)~coarse-to-fine exploration of assets and drafts (\texttt{describe\_media}, \texttt{locate\_event}, \texttt{inspect\_segment}, \texttt{transcribe\_speech}); (ii)~verification of requirements quoted from the brief, with a deterministic check that the evidence covers every required modality (\texttt{check\_requirement}); (iii)~comparison of two media on separate clocks, against a reference or across revisions to confirm a repair (\texttt{compare\_media}); and (iv)~local probes of decoding, loudness, silence, frozen frames, and loop seams that need no API call (\texttt{probe\_media}). A lightweight staged protocol (\texttt{plan\_production}) links quoted constraints in the brief to ordered production beats before any code is written, and has the agent test its riskiest decision on a minimal preview before moving on to assembly, targeted repair, and delivery.

Because the primitives operate on standard image, audio, and video containers, VibeVideoAgent needs no adapters for particular rendering engines: the same seven primitives serve 3D WebGPU/Canvas scenes, React/Remotion interfaces, Cairo and Manim graphics, SVG game simulations, and multi-track TTS/FFmpeg audio (Appendix~\ref{sec:app_prompts}). We expose them through the Model Context Protocol (MCP)~\citep{anthropic2024mcp} to the native agentic coding CLIs, Claude Code and Codex CLI, with Gemini~3.8 Flash as a fixed perception model, and compare each coding model (Claude Opus~5.5, Claude Sonnet~5.5, and GPT-6 Astra) with its own native agent on all 239 VibeVideoBench tasks. Our contributions are:
\begin{itemize}
  \item \textbf{A perceptual framework for in-production video authoring.} A training-free architecture that keeps dense audiovisual inspection out of the long-horizon coding context, so that agents can check early previews and repair their programs iteratively across heterogeneous rendering pipelines without engine-specific adapters.
  \item \textbf{Multi-clock perception and verification primitives.} A unified multi-clock draft representation and seven primitives for coarse-to-fine exploration, modality-gated requirement checks, comparison with references and earlier revisions on separate clocks, and local signal probes, used under a lightweight staged production protocol.
  \item \textbf{Evaluation with three coding models.} On all 239 tasks, VibeVideoAgent improves L3 over each native agent: $+0.40$ for Claude Opus~5.5 (with $99.35\%$ L2), $+0.24$ for GPT-6 Astra (with $99.51\%$ L1), and $+0.13$ for Claude Sonnet~5.5. Ablations and trajectory telemetry isolate each component and show how the frequency of verification and the granularity of edits shape the balance between L2 and L3.
\end{itemize}

\section{Related Work}
\label{sec:related}

\paragraph{Code-driven visual and video generation.}
Visual programming showed that language models can compose vision modules through executable code~\citep{gupta2023visprog,suris2023vipergpt}. Code generation now produces charts and plots~\citep{yang2024matplotagent,yang2025chartmimic}, web pages and interactive artifacts~\citep{si2025design2code,zhang2025artifactsbench,zhang2026animweb}, and animations~\citep{tseng2024keyframer,liu2024logomotion}, and MoVer~\citep{ma2025mover} verifies generated motion graphics against logical specifications of their motion. For video, TheoremExplainAgent~\citep{ku2025theoremexplainagent} and Code2Video~\citep{chen2026code2video} render educational Manim videos with storyboarding and layout critique, Paper2Video~\citep{zhu2025paper2video} and PresentAgent~\citep{shi2025presentagent} turn documents into narrated presentations, Lab2Sim~\citep{hei2026lab2sim} turns experiment footage into simulation code, and WavJourney~\citep{liu2023wavjourney} composes audio from LLM-written scripts. SenseNova~6.8 Code2Video~\citep{sensenova2026code2video} is an open-weight model that renders short videos from the code it writes, OpenMotion~\citep{openmotion2026} has coding agents rebuild reference videos as editable HTML, and VibeVideoBench~\citep{diao2026vibevideobench} benchmarks the task in 12 categories. Most of these systems are tied to one medium or engine, whereas VibeVideoAgent offers one perceptual interface across graphics libraries, browsers, simulation loops, and multi-track audio.

\paragraph{Agentic 3D scene synthesis and previsualization.}
Code also gives explicit control over 3D geometry, layout, and camera~\citep{sun2023threedgpt,yang2024holodeck,du2024blenderllm}. Several 3D systems check rendered views with a VLM, for critique~\citep{hu2024scenecraft}, edit search~\citep{huang2024blenderalchemy}, visual question answering on CAD objects~\citep{alrashedy2024cadcodeverify}, or multi-agent refinement~\citep{lu2025ll3m}, but they look at still renders of a single program state. CineRecon~\citep{wang2026cinerecon} and VibePrevisBench~\citep{chen2026vibeprevisbench} evaluate multi-shot 4D reconstruction and previsualization, and find that agents fail when they cannot reconcile scene logic in code with what appears on screen. VibeVideoAgent addresses this screen-space gap over time, with reference-to-draft comparison, preview checks, and verification across revisions, and without engine-specific scene-graph extractors.

\paragraph{Feedback-driven agents.}
Agents improve their outputs through self-refinement and reflection~\citep{madaan2023selfrefine,shinn2023reflexion}, execution feedback on code~\citep{chen2024selfdebug,wang2024codeact}, and agent--computer interfaces~\citep{yao2023react,yang2024sweagent,wang2025openhands}. With visual feedback, Visual Sketchpad~\citep{hu2024sketchpad} lets multimodal models inspect their own sketches, Vision-as-a-Tool~\citep{yang2026vaat} gives coding agents inspection tools for static HTML, SVG, and 3D assets, and PhysLoop~\citep{chen2026physloop} refines pixel-space video generators with physics-simulation feedback. Generated video is usually scored after the fact, by benchmark suites~\citep{huang2024vbench} or multimodal LLM judges~\citep{chen2024mllmjudge}; VibeVideoAgent instead uses a multimodal model during production, on bounded excerpts and under deterministic validation.

\paragraph{Agentic video understanding.}
For long videos~\citep{wang2025lvbench,fu2025videomme,comanici2025gemini25}, agents decide where to look: they compose models through language~\citep{zeng2023socratic}, retrieve frames or query a memory~\citep{wang2024videoagent,fan2024videoagent}, search images for small details~\citep{wu2024vstar}, or search long videos with tools~\citep{zhang2025dvd,lin2026videoseek}. Sprout~\citep{chen2026sprout} replaces accumulated video inputs with a textual memory tree built while answering, and VideoACE~\citep{zhang2026videoace} tests whether agents adapt temporal coverage, sampling density, and spatial detail to fast actions. Our work builds on VideoNexus~\citep{long2026videonexus}, which showed that a planner can direct perception run in independent contexts and reason over compact, temporally grounded reports, with no visual tokens in its own context. These methods answer questions about a fixed, finished video. Authoring is different: the media are mutable, incomplete, and spread over source, preview, and film clocks; claims often span audio and vision; and each code edit shifts event timing. VibeVideoAgent therefore adds multi-clock addressing, modality-gated verification, comparison across revisions on separate clocks, and deterministic signal probes.

\section{Method}
\label{sec:method}

\begin{figure*}[t]
  \centering
  \includegraphics[width=\linewidth]{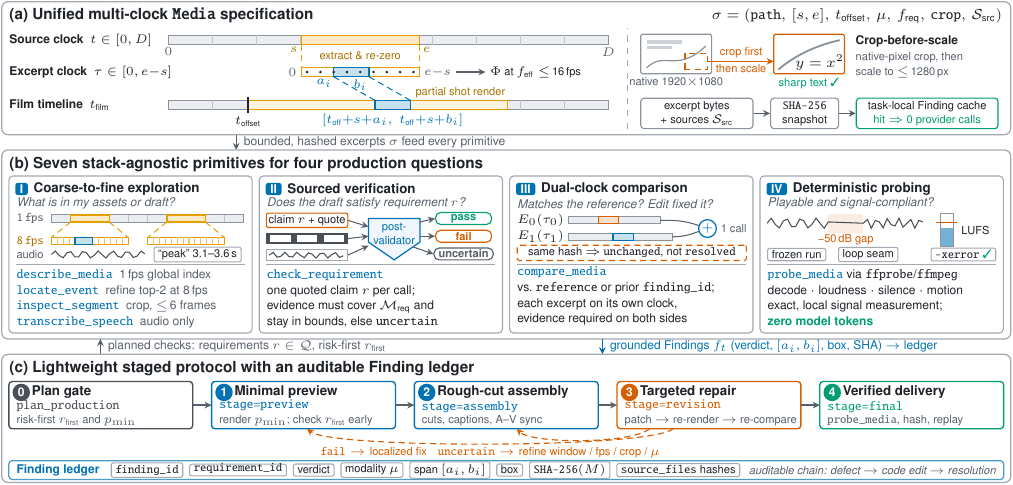}
  \caption{\textbf{Architecture of VibeVideoAgent.} \textbf{(a)}~A three-clock \texttt{Media} specification $\sigma$ addresses assets, stills, partial previews, and full renders, with crop-before-scale magnification and source-file hashing. \textbf{(b)}~Seven stack-agnostic primitives explore media from coarse to fine, verify sourced requirements across modalities, compare two media on separate clocks, and probe signals without model calls, returning grounded Findings $f_t$ to an auditable ledger. \textbf{(c)}~A staged protocol gates planning and then guides preview, assembly, repair, and delivery checks.}
  \label{fig:architecture}
\end{figure*}

VibeVideoAgent lets a general-purpose coding agent check its own video drafts without changing model weights or committing to a rendering library (Figure~\ref{fig:architecture}).

\subsection{Decoupled Perceptual Loop for Video Authoring}
\label{sec:formulation}

A coding agent must produce a target video and its editable source repository from a public brief $\mathcal{B}$ and read-only input assets $\mathcal{A} = \{A_1, \dots, A_m\}$ (documents, data tables, images, audio, or reference videos). Over steps $t = 1, \dots, T$, it maintains a mutable codebase $P_t$ and a reasoning context $C_t$, and it may run rendering code to produce an intermediate or final artifact
\begin{equation}
  M_t = \mathcal{R}(P_t, \omega_t),
  \label{eq:render}
\end{equation}
where $\omega_t$ is the render scope, for example a single design frame, a $3$-second preview of Shot~2, or a full rough cut with mixed audio. In a \emph{coupled} multimodal loop, inspecting $M_t$ over a window of duration $\Delta t$ at frame rate $f$ appends raw tokens to $C_t$:
\begin{equation}
  C_{t+1} = C_t \oplus a_t \oplus o_t^{\text{coupled}}, \quad |o_t^{\text{coupled}}| \approx \Delta t\,\bigl(f\,\tau_v + \tau_a\bigr),
  \label{eq:coupled}
\end{equation}
where $\tau_v$ and $\tau_a$ are the token costs per visual frame and per second of audio. Across repeated previews, multi-shot debugging, and synchronization checks, these tokens pile up next to hundreds of lines of rendering code, which raises attention cost and hurts long-horizon programming.

VibeVideoAgent instead uses a \emph{decoupled} loop. Raw media never enter $C_t$. Each perceptual query goes to a stateless worker $\Phi$ in an isolated, short-lived context (or to a local deterministic analyzer $\Psi$) and returns a compact, structured \emph{Finding}
\begin{equation}
  f_t = \Phi\bigl(\mathcal{E}(M_t, \sigma_t),\, q_t\bigr), \qquad |f_t| \ll |\mathcal{E}(M_t, \sigma_t)|,
  \label{eq:decoupled}
\end{equation}
where $\mathcal{E}(M_t, \sigma_t)$ extracts a bounded excerpt under the sampling specification $\sigma_t$ and $q_t$ is a focused perception or verification query. A Finding $f_t = (\texttt{id}, \texttt{verdict}, \texttt{summary}, \mathcal{V})$ contains a typed verdict, a short summary, and evidence $\mathcal{V} = \{(m_i, [a_i, b_i], \mu_i, b_i^{\text{box}}, \text{text}_i)\}$ localized by time interval, modality $\mu_i \in \{\texttt{visual}, \texttt{audio}\}$, and normalized box $b_i^{\text{box}} \in [0,1]^4$. Frames and audio stay inside the short-lived call and are discarded afterward, so $C_t$ stays small and focused on code.

\subsection{Unified Multi-Clock Draft Representation}
\label{sec:multiclock}

Production media differ from the single video of a question-answering task: they exist at several stages of completion and run on different clocks. A creator may at once handle a $45$-second raw clip in \texttt{inputs/}, a static $1920{\times}1080$ character sheet, a $4$-second preview rendered only for Scene~3 (which starts at $12.0\,$s in the film), and a $30$-second rough cut. VibeVideoAgent addresses all of them with one \texttt{Media} specification,
\begin{equation}
  \sigma = \bigl(\texttt{path},\, [s, e],\, t_{\text{offset}},\, \mu,\, f_{\text{req}},\, \texttt{crop},\, \mathcal{S}_{\text{src}}\bigr),
  \label{eq:mediaspec}
\end{equation}
which reconciles three coordinate systems and enforces four invariants (Figure~\ref{fig:architecture}a).

\paragraph{Three-clock temporal mapping.}
Every interval $[s, e]$ is given in the \emph{source clock} of \texttt{path} ($s=e=0$ for stills). The clip shown to the perception model starts at zero on its own \emph{excerpt clock} $\tau \in [0, e-s]$, which keeps the model from hallucinating timestamps on pre-cut clips. Each evidence span $[a_i, b_i]$ is mapped back deterministically to the source range $[s+a_i,\, s+b_i]$ and, when $t_{\text{offset}}$ is given for a partial render, to the \emph{film timeline} $[t_{\text{offset}} + s + a_i,\; t_{\text{offset}} + s + b_i]$.

\paragraph{Crop-before-scale magnification.}
Downscaling a 1080p or 4K frame blurs small labels, subscripts, hand gestures, and 3D contact points. Given $\texttt{crop} = [x, y, w, h]$ in source pixels, the extractor first crops the frame at native resolution and then scales the region (up to $1280\,$px for video and $2048\,$px for stills); boxes are reported relative to the cropped view.

\paragraph{Adaptive sampling and modality routing.}
With $\mu = \texttt{auto}$, the service inspects the container and selects \texttt{audiovisual}, \texttt{visual}, or \texttt{audio}, sending both streams only when a query needs sight and sound together. The effective video sampling rate is
\begin{equation}
  f_{\text{eff}} = \min\!\left(f_{\text{req}},\; f_{\max},\; \frac{B_{\text{frames}}}{e - s},\; f_{\text{native}}\right),
  \label{eq:fps}
\end{equation}
which respects the per-call frame cap $B_{\text{frames}}$, the ceiling $f_{\max}=16\,$fps, and the native rate $f_{\text{native}}$.

\paragraph{Immutable snapshots and provenance caching.}
Drafts are overwritten from one render to the next, so the extractor checks that the file is stable during sampling, stores an immutable excerpt identified by its SHA-256 hash, and records SHA-256 digests of the declared editable source files $\mathcal{S}_{\text{src}}$ (e.g., \texttt{scene.py}). Repeated queries on unchanged bytes with unchanged parameters are answered from a task-local Finding cache without calling the provider.

\subsection{Seven Perception and Verification Primitives}
\label{sec:primitives}

Instead of one open-ended prompt that asks a VLM about the video, VibeVideoAgent provides seven primitives, independent of the rendering stack and organized around four questions an author faces during production (Figure~\ref{fig:architecture}b; Appendix~\ref{sec:app_primitives} lists verdict sets and defaults).

\paragraph{I. What is in my assets or current draft?}
Dense uniform sampling wastes budget and a single sparse pass misses short transitions, so exploration proceeds from coarse to fine. \texttt{describe\_media} returns a $1\,$fps index of scenes, text, and sounds (or the layout of a still) and is told not to treat an event missed at sparse sampling as absent. \texttt{locate\_event} searches for a queried object, action, or sound; on sparse scans ($f_{\text{eff}} < 8\,$fps) it automatically re-queries up to two candidate windows at $8\,$fps to pin down onsets and cuts. \texttt{inspect\_segment} answers a diagnostic question about a chosen window or pixel \texttt{crop} at $4\,$fps and can return up to six evidence frames to the coder. \texttt{transcribe\_speech} uploads only the audio and returns timestamped verbatim utterances for cutting dialogue or aligning animation with narration.

\paragraph{II. Does the draft satisfy a requirement?}
Open-ended VLM critiques tend to comment on aesthetics nobody asked about and to overlook subtle constraints, especially on cross-modal timing. \texttt{check\_requirement} instead tests one explicit observable claim ($2\,$fps by default) together with its quoted \texttt{source} and required modalities $\mathcal{M}_{\text{req}} \subseteq \{\texttt{visual}, \texttt{audio}\}$. The claim is passed directly or looked up by \texttt{requirement\_id} in the production plan, and paths inside \texttt{inputs/} are rejected so that verification always targets a produced draft. A deterministic post-validator maps the model's raw response $(\hat{v}, \mathcal{V})$ to
\begin{equation}
  v_{\text{final}} =
  \begin{cases}
    \texttt{uncertain} & \text{if } \mathcal{V} = \emptyset \text{ or } \mathcal{M}_{\text{req}} \not\subseteq \mu(\mathcal{V}), \\
    \hat{v} & \text{otherwise},
  \end{cases}
  \label{eq:check_gate}
\end{equation}
where $\hat{v} \in \{\texttt{pass}, \texttt{fail}, \texttt{uncertain}\}$ and $\mu(\mathcal{V})$ is the set of modalities cited in $\mathcal{V}$, computed after discarding evidence whose span exceeds the excerpt or whose box violates $0 \le x_1 < x_2 \le 1$, $0 \le y_1 < y_2 \le 1$. For instance, the claim \emph{``the peak marker is visible when the narrator says `peak'\,''} requires $\mathcal{M}_{\text{req}} = \{\texttt{visual}, \texttt{audio}\}$. If the excerpt has no audio or the worker cites only visual evidence, the verdict becomes \texttt{uncertain}, and the agent must supply both streams or widen the window before it can declare success.

\paragraph{III. Does it match the reference, and did my edit fix the issue?}
Reference matching and repair both compare two artifacts whose timings rarely coincide: fixing a delay or trimming a voiceover shifts later timestamps, and a reconstruction only approximates the pacing of its reference. \texttt{compare\_media} places two excerpts $[E_0, E_1]$ in a single call, each on its \emph{own} zero-based clock, and requires evidence from both (otherwise \texttt{uncertain}). In \emph{reference mode}, it compares a reference image or clip $E_0$ with a draft $E_1$ along given \texttt{focus} attributes (layout, camera path, palette, or choreography) and returns \texttt{changed}, \texttt{unchanged}, or \texttt{uncertain}. In \emph{revision mode}, it takes the \texttt{finding\_id} of an earlier defect, retrieves the immutable excerpt stored at that failed check (after verifying its hash), and pairs it with the new render. It inherits the parent Finding's required modalities and asks whether the issue is \texttt{resolved} or \texttt{persists}, reporting any regressions nearby. A hash guard turns \texttt{resolved} into \texttt{unchanged} whenever $\text{SHA256}(M_0) = \text{SHA256}(M_1)$, which catches edits that were never re-rendered, and the ledger diffs the hashes of the declared \texttt{source\_files}, so each defect can be traced to a code edit and its resolution.

\paragraph{IV. Is the file playable and signal-compliant?}
Multimodal models cannot detect a corrupt packet at frame $587$, measure loudness in LUFS, flag a $0.3\,$s gap of dead air, or verify a pixel-exact loop from sparse samples. \texttt{probe\_media} runs locally with \texttt{ffprobe}/\texttt{ffmpeg}, makes no API calls, and reports stream metadata plus four measurements: a strict full-stream \texttt{decode} (\texttt{ffmpeg -xerror}); \texttt{loudness} (integrated loudness, loudness range, and true peak); \texttt{silence} (intervals below $-50\,$dB lasting at least $0.25\,$s); and \texttt{motion} (exact frame count, the longest run of frozen frames, and the RGB difference between the first and last frames for loop seams).

\subsection{Staged Production Protocol}
\label{sec:protocol}

Good perception tools answer \emph{how} to inspect media but not \emph{when}. Left alone, an agent may write an entire multi-minute project before its first check, or never check its riskiest requirement. VibeVideoAgent therefore pairs the primitives with a lightweight staged protocol built around one local planning ledger (Figure~\ref{fig:architecture}c).

\paragraph{Stage 0: plan gate (\texttt{plan\_production}).}
After reading the brief and listing \texttt{inputs/}, and \emph{before} any code edit, render, or media call (a pre-tool runtime gate enforces this), the agent records a plan $(\mathcal{G}, \mathcal{D}, \mathcal{Q}, \mathcal{B}_{\text{beats}}, \mathcal{C}_{\text{choice}}, r_{\text{first}}, p_{\text{min}})$. Each requirement $r \in \mathcal{Q}$ must quote the brief verbatim (checked programmatically against $\mathcal{B}$), state its observable phenomenon and required modalities, and link to at least one ordered beat in $\mathcal{B}_{\text{beats}}$. This separates what the user asked for from the creative choices $\mathcal{C}_{\text{choice}}$ left to the agent. The agent also names the highest-risk requirement $r_{\text{first}}$, whose failure would force the most rework, and the smallest render $p_{\text{min}}$ that can test it.

\paragraph{Stages 1--4: preview, assembly, repair, delivery.}
In the \emph{minimal preview} stage (\texttt{stage=preview}), the agent explores unfamiliar assets, renders $p_{\text{min}}$ (a still for typography or character design, a short clip for motion or camera, or an audiovisual slice for narration sync), and checks $r_{\text{first}}$ before building the remaining beats. In \emph{rough-cut assembly} (\texttt{stage=assembly}), it verifies cross-shot properties such as ordering, transitions, caption legibility, and speech-to-graphic sync on cheap rough cuts. In \emph{targeted repair} (\texttt{stage=revision}), an \texttt{uncertain} verdict leads the agent to refine the window, frame rate, crop, or modality instead of guessing an edit, and a \texttt{fail} verdict shows where to patch; the agent then re-renders the affected slice and calls \texttt{compare\_media(finding\_id, media)} to confirm \texttt{resolved} and check the shifted neighbors. In \emph{verified delivery} (\texttt{stage=final}), it runs \texttt{probe\_media} on the encoded MP4, re-verifies requirements that compositing or mixing could break, registers the immutable output with its source files and a replayable render command, and submits.

\section{Experiments}
\label{sec:experiments}

\begin{table*}[t]
\centering
\caption{\textbf{Main results on VibeVideoBench (239 tasks, 12 categories).} Each coding model's official \textbf{Native Agent} vs.\ \textbf{+ VibeVideoAgent} (Gemini~3.8 Flash perception) in identical sandboxes. Scores average the official Gemini~3.8 Flash and Gemini~4 Argon judges (L1/L2 in \%; L3 rated 1--5); bold marks the better configuration.}
\label{tab:vvb_main}
\small
\setlength{\tabcolsep}{5pt}
\begin{tabular}{@{}ll cc cccccc@{}}
\toprule
\textbf{Coding backbone} & \textbf{Configuration} & \textbf{L1 Spec} $\uparrow$ & \textbf{L2 Rubric} $\uparrow$ & \textbf{VQ} $\uparrow$ & \textbf{VA} $\uparrow$ & \textbf{CL} $\uparrow$ & \textbf{TM} $\uparrow$ & \textbf{AQ} $\uparrow$ & \textbf{L3 Mean} $\uparrow$ \\
\midrule
\multirow{2}{*}{Claude Opus 5.5}
  & Native Agent & \textbf{99.92} & 98.87 & 4.04 & 4.08 & 4.09 & 4.03 & 3.89 & 4.03 \\
  & \textbf{+ VibeVideoAgent} & 98.59 & \textbf{99.35} & \textbf{4.37} & \textbf{4.43} & \textbf{4.46} & \textbf{4.42} & \textbf{4.45} & \textbf{4.43} \\
\midrule
\multirow{2}{*}{Claude Sonnet 5.5}
  & Native Agent & \textbf{99.18} & \textbf{96.36} & 3.97 & 4.02 & 4.03 & 3.96 & 3.85 & 3.97 \\
  & \textbf{+ VibeVideoAgent} & 95.75 & 94.61 & \textbf{3.98} & \textbf{4.05} & \textbf{4.15} & \textbf{4.07} & \textbf{4.25} & \textbf{4.10} \\
\midrule
\multirow{2}{*}{GPT-6 Astra}
  & Native Agent & 98.93 & 98.67 & 4.07 & 4.09 & 4.08 & 3.98 & 3.94 & 4.03 \\
  & \textbf{+ VibeVideoAgent} & \textbf{99.51} & \textbf{99.15} & \textbf{4.26} & \textbf{4.31} & \textbf{4.29} & \textbf{4.20} & \textbf{4.26} & \textbf{4.27} \\
\bottomrule
\end{tabular}
\end{table*}

\subsection{Experimental Setup}
\label{sec:setup}

\paragraph{Benchmark and metrics.}
We evaluate on VibeVideoBench~\citep{diao2026vibevideobench}, which has 239 tasks in 12 English and Chinese production categories, such as data storytelling, paper explainers, motion graphics, interactive 3D demos, game simulations, narrative animations, reference-guided reconstructions, and music-driven edits. Agents receive a natural-language brief and read-only assets, cannot call pixel-generation APIs, and must write rendering software that delivers a playable MP4 and an editable source repository. Two official multimodal judges, Gemini~3.8 Flash and Gemini~4 Argon, score every delivery. We report their equal-weight average and per-judge results for \textbf{L1} (compliance with technical delivery specifications), \textbf{L2} (pass rate over thousands of binary rubric items), and \textbf{L3} (five perceptual dimensions rated 1--5: Visual Quality \emph{VQ}, Visual Aesthetics \emph{VA}, Content Logic \emph{CL}, Timing \& Motion \emph{TM}, and Audio Quality / A--V Sync \emph{AQ}, and their mean).

\paragraph{Controlled single-variable comparison.}
Each coding model runs in its native CLI: Claude Code (\texttt{claude -p}, effort \texttt{max}) for Claude Opus~5.5 and Claude Sonnet~5.5, and Codex CLI for GPT-6 Astra. All runs use isolated 16-vCPU Linux namespaces with identical rendering libraries and read-only inputs (\texttt{0444}). Every \textbf{+ VibeVideoAgent} run registers the same MCP server and a \texttt{PreToolUse} hook that blocks direct reads of raw media, sub-agents, and ad-hoc external API calls, so all perception goes through the seven audited primitives; wall time, provider tokens, and retries are logged. The perception model is fixed to Gemini~3.8 Flash (\texttt{gemini-3.8-flash}). Each \textbf{+ VibeVideoAgent} delivery is paired with the model's official \textbf{Native Agent} baseline on the same task, and component ablations use Claude Opus~5.5 on a 36-task diagnostic subset (Section~\ref{sec:ablations}). Hidden rubrics and judges stay outside the authoring filesystem and run only after the output MP4 and source snapshot are frozen by SHA-256.

\subsection{Main Results}
\label{sec:main_results}

\paragraph{VibeVideoAgent improves every coding model.}
VibeVideoAgent raises the dual-judge L3 Mean of all three models (Table~\ref{tab:vvb_main}), by $0.40$ for Claude Opus~5.5, $0.24$ for GPT-6 Astra, and $0.13$ for Claude Sonnet~5.5.
\emph{(i)~The largest gains are temporal and cross-modal.} \emph{AQ} improves the most for every model ($+0.56$ for Opus, $+0.40$ for Sonnet, $+0.32$ for Astra), followed by \emph{TM} and \emph{CL}. Native agents often write valid FFmpeg or TTS pipelines whose narration drifts behind transitions, is cut off at shot boundaries, or leaves dead air; modality-gated \texttt{check\_requirement}, \texttt{transcribe\_speech}, and \texttt{probe\_media} expose these failures while the video is still in production.
\emph{(ii)~The gains are not tied to one judge.} Gemini~3.8 Flash is both the perception model and one of the judges. Under this judge, L3 rises from $4.14$ to $4.60$ for Opus ($124$ task wins, $11$ losses), from $4.10$ to $4.43$ for Sonnet, and from $4.11$ to $4.59$ for Astra ($121$ wins, $16$ losses). Under the independent and stricter Argon judge, Opus still gains $0.80\,$pp L2 ($98.36\% \to 99.15\%$) and $0.33$ L3 ($3.91 \to 4.25$; $132$ wins, $37$ losses), and Astra raises L2 from $98.38\%$ to $99.00\%$.
\emph{(iii)~The interface transfers across CLI runtimes.} With VibeVideoAgent, Opus reaches the highest L2 and L3 in Table~\ref{tab:vvb_main} ($4.37$--$4.46$ on all five dimensions) and Astra the highest L1, so the same MCP interface works in both Claude Code and Codex CLI.

\begin{table*}[t]
\centering
\caption{\textbf{Context decoupling and repair-funnel telemetry on VibeVideoBench (per task).} Worker calls, Findings, and sampled media stay in short-lived perception contexts; \emph{Coding Out Tok} counts the coding model's output tokens; the last three columns track staged verification and repair convergence.}
\label{tab:funnel}
\small
\begin{threeparttable}
\begin{tabular}{@{}l cccccc@{}}
\toprule
\textbf{Coding backbone} & \makecell{\textbf{Calls /}\\\textbf{Findings}} & \makecell{\textbf{Sampled Media}\\\textbf{(Fr.\ / Audio)}} & \makecell{\textbf{Coding Out}\\\textbf{Tok / Task}} & \makecell{\textbf{Preview Gate}\\\textbf{Checked (\%)}} & \makecell{\textbf{Failed Checks}\\\textbf{Edited (\%)}} & \makecell{\textbf{Dual-Clock}\\\textbf{Resolved (\%)}} \\
\midrule
Claude Opus 5.5   & 12.0 / 6.2  & 386 / 171\,s     & 147.9k & 93.9$^{\ddagger}$ & 77.4 & 58.3 \\
Claude Sonnet 5.5 & 12.2 / 7.8  & 303 / 138\,s     & 206.8k & 84.3              & 75.3 & 42.5 \\
GPT-6 Astra       & 50.6 / 23.1 & 1{,}297 / 660\,s & 179.3k & 100.0             & 84.2 & 65.6 \\
\bottomrule
\end{tabular}
\begin{tablenotes}[flushleft]\footnotesize
\item[$\ddagger$] $93.9\%$ ($138/147$) on trajectories with accepted worker Findings ($57.7\%$ over all $239$ tasks); the upfront \texttt{plan\_production} step succeeds on $100.0\%$ of tasks for all three models.
\end{tablenotes}
\end{threeparttable}
\end{table*}

\paragraph{Verification frequency and edit granularity.}
The telemetry in Table~\ref{tab:funnel} reveals three distinct behaviors.
\emph{Claude Opus~5.5 checks selectively and makes small patches} ($+0.48\,$pp L2, $+0.40$ L3). It edits after $77.4\%$ ($171/221$) of failed checks and resolves $58.3\%$ ($88/151$) of dual-clock rechecks with the least coding output of the three models. Under the Argon judge, its narrative animations (\texttt{N*}) reach $100.00\%$ L2 on all $26$ tasks, and its paper explainers (\texttt{PA*}) gain $3.03\,$pp L2.
\emph{GPT-6 Astra checks often and patches over many revisions} ($+0.58\,$pp L1, $+0.48\,$pp L2, $+0.24$ L3). It makes $50.6$ worker calls per task and has the highest dual-clock resolution rate ($65.6\%$, $419/639$); its output tokens are spread over $7.0$ versions ($\approx 25.7\text{k}$ each) as local diffs that fix missed rubric items and leave working code alone.
\emph{Claude Sonnet~5.5 finds defects but repairs them coarsely} ($+0.13$ L3, $-1.75\,$pp L2). It flags $312$ \texttt{fail} and $214$ \texttt{uncertain} verdicts, and acting on them raises \emph{AQ} by $0.40$. However, it produces the most coding output and resolves only $42.5\%$ ($34/80$) of its revision comparisons, because it often rewrites whole rendering scripts instead of patching single functions. The rewrites improve pacing and sync but lose secondary static details ($95.87\% \to 92.37\%$ Argon L2).
Perceptual feedback thus improves rubric compliance and perceptual quality together only when the agent keeps its edits local.

\subsection{Ablations and In-Production Analysis}
\label{sec:ablations}

\paragraph{Treatment-blind diagnostic subset.}
More than $80\%$ of tasks already reach $100\%$ L2 with Native Opus~5.5, so a random subset leaves little room to separate variants, while selecting tasks by the gains of VibeVideoAgent would bias the comparison through selection and regression to the mean. We therefore build a deterministic subset of $36$ tasks ($15\%$ of the benchmark) from native-baseline diagnostics and public task metadata alone, before running any ablation. Its four strata cover rubric bottlenecks ($12$ tasks), reference-grounded assets ($8$), audiovisual and temporal bottlenecks ($10$), and median-baseline guardrails ($6$); see Appendix~\ref{sec:app_subset}.

\begin{table*}[t]
\centering
\caption{\textbf{Component ablations on the 36-task treatment-blind, capability-stratified subset} (Claude Opus~5.5, Gemini~3.8 Flash perception, dual-judge averages). \textbf{(a)}~Removing individual primitives. \textbf{(b)}~Removing the staged production protocol while keeping all seven primitives.}
\label{tab:ablations}
\small
\setlength{\tabcolsep}{4.5pt}
\begin{tabular}{@{}l cc cccccc@{}}
\toprule
\textbf{Ablation variant} & \textbf{L1 Spec} $\uparrow$ & \textbf{L2 Rubric} $\uparrow$ & \textbf{VQ} $\uparrow$ & \textbf{VA} $\uparrow$ & \textbf{CL} $\uparrow$ & \textbf{TM} $\uparrow$ & \textbf{AQ} $\uparrow$ & \textbf{L3 Mean} $\uparrow$ \\
\midrule
\multicolumn{9}{@{}l}{\emph{(a) Verification and signal-probing primitives}} \\
Native coding agent (no perception tools) & \textbf{100.00} & 96.06 & 3.87 & 3.90 & 3.89 & 3.76 & 3.80 & 3.84 \\
VibeVideoAgent w/o \texttt{check\_requirement} & 98.43 & 98.35 & 4.27 & 4.26 & 4.31 & 4.26 & 4.33 & 4.29 \\
VibeVideoAgent w/o \texttt{compare\_media} & 97.38 & 99.11 & 4.20 & 4.37 & 4.39 & 4.26 & 4.28 & 4.30 \\
VibeVideoAgent w/o \texttt{probe\_media} & 95.81 & 99.03 & 4.26 & 4.33 & 4.40 & 4.31 & 4.36 & 4.33 \\
\midrule
\multicolumn{9}{@{}l}{\emph{(b) Production orchestration}} \\
Tools-Only (w/o plan gate \& preview gate) & 98.95 & 99.03 & 4.26 & 4.27 & 4.43 & 4.29 & 4.34 & 4.32 \\
\textbf{VibeVideoAgent (Full)} & 98.43 & \textbf{99.50} & \textbf{4.34} & \textbf{4.41} & \textbf{4.46} & \textbf{4.39} & \textbf{4.41} & \textbf{4.40} \\
\bottomrule
\end{tabular}
\end{table*}

\paragraph{Component ablations.}
The full system has the best L2 ($3.44\,$pp above Native; $94.28\% \to 99.11\%$ under Argon) and the best score on every L3 dimension ($+0.56$ L3 Mean; Table~\ref{tab:ablations}), and each ablation hurts a different aspect of the output.
\emph{Without \texttt{check\_requirement}}, rubric precision and coherence suffer most: L2 drops by $1.15\,$pp ($1.73\,$pp under Argon), and \emph{CL} and \emph{VA} each fall by $0.15$. Ad-hoc inspection without quoted claims and required audio and visual evidence misses subtle constraints and accepts guesses from a single modality.
\emph{Without \texttt{compare\_media}}, repairs and timing suffer: L3 falls to $4.30$, with the largest drops on \emph{VQ}, \emph{TM}, and \emph{AQ}. The share of failed checks followed by an edit drops from $94.0\%$ ($47/50$) to $78.3\%$ ($65/83$), and motion fixes, reference mismatches, and audio shifts introduced by edits go unverified.
\emph{Without \texttt{probe\_media}}, playability suffers: L1 drops the most ($-2.62\,$pp), and L3 falls by $0.16$ under Argon, because sparse frame sampling cannot replace exact decoding, LUFS loudness, $-50\,$dB silence, and frozen-frame analysis.
\emph{Without the staged protocol} (Tools-Only, which keeps all seven primitives but drops \texttt{plan\_production} and the preview gate), L2 falls by $0.47\,$pp ($1.05\,$pp under Argon) and L3 to $4.32$. Preview-gate adherence drops from $97.2\%$ to $0.0\%$, and dual-clock resolution falls from $70.6\%$ ($24/34$) to $52.2\%$ ($12/23$), as untested layout and rigging choices turn into expensive late rewrites.

\paragraph{Context decoupling.}
Per task, perception workers process $386$ frames and $171\,$s of audio for Opus~5.5, $303$ frames and $138\,$s ($75.2\text{k}$ prompt tokens) for Sonnet~5.5, and $1{,}297$ frames and $660\,$s ($335.0\text{k}$ prompt tokens) for GPT-6 Astra, all outside the coding session, while the Findings stored in \texttt{state.json} average only $0.3\text{k}$--$1.4\text{k}$ tokens. Replaying the same media through a coding history of 30--50 turns would cost tens of millions of input tokens per task. On an eight-task suite with exact single-turn response sizes (Appendix~\ref{sec:app_pilot}), the returned Findings are $6.5\times$ smaller than the perceived media ($12.0\times$ on the long-form explainer \texttt{PA02}), and L2 rises from $90.2\%$ to $96.0\%$ (Flash) and from $85.0\%$ to $94.5\%$ (Argon).

\paragraph{Qualitative trajectory.}
Appendix~\ref{sec:app_case} (Figures~\ref{fig:trajectory} and~\ref{fig:hero}) walks through a recorded run on \texttt{W0448}. The agent finds three staging bugs in a render that exited cleanly, confirms both patches with dual-clock \texttt{compare\_media}, and returns \texttt{uncertain} where evidence is incomplete; Flash-judged L2 rises from $76.0\%$ to $96.0\%$.

\section{Limitations}
\label{sec:limitations}

Perception samples bounded frame rates ($\le 16\,$fps) and selected intervals, so sub-frame glitches or windows that were never inspected can go unnoticed, even with $8\,$fps refinement and full-stream decoding. VibeVideoAgent reports unobserved spans and missing modalities as \texttt{uncertain}, and the delivery check certifies that the file plays correctly; it does not certify the absence of semantic defects. Comparing hash-bound clips from before and after an edit shows that a rendered symptom changed after a declared edit. This is an observation and not a causal proof over program branches: an edit to a shared animation helper can affect beats that were never inspected if the agent skips neighbor checks. Finally, the external perception model (Gemini~3.8 Flash) can misread dense overlapping text or fine 3D depth ordering unless the agent requests a \texttt{crop} or evidence frames, intermediate renders add compute and latency, and the behavior of commercial APIs may change with provider updates.

\section{Conclusion}
\label{sec:conclusion}

Code-driven video generation gives agents precise control over layout, motion, sound, and camera, provided they can see and check what their code renders. VibeVideoAgent provides this feedback without filling the coding context with media: a multi-clock media specification and seven primitives, used under a lightweight staged protocol, let the agent catch defects early and verify its repairs across heterogeneous rendering stacks. On all 239 VibeVideoBench tasks, it improves holistic audiovisual quality for three frontier coding models (up to $+0.40$ L3 and $+0.56$ on A--V sync) while raw media stay outside the coding session, and when the agent edits locally, rubric adherence and perceptual quality improve together. Future work includes guidance on edit locality for smaller coding models, adaptive scheduling of preview fidelity, open-weight perception models, and dedicated engines for 4D previsualization~\citep{chen2026vibeprevisbench}.

\setlength{\bibsep}{3.0pt plus 1pt}
\bibliographystyle{plainnat}
\bibliography{references}

\clearpage
\appendix
\section{Primitive and Protocol Details}
\label{sec:app_primitives}

Table~\ref{tab:primitives} summarizes the seven primitives of Section~\ref{sec:primitives} with their verdict sets and default settings. Appendices~\ref{sec:app_schemas}--\ref{sec:app_isolation} give the exact tool contracts, the deterministic validation rules, the workflow guidance, and the runtime isolation; Appendices~\ref{sec:app_subset}--\ref{sec:app_case} describe how the ablation subset was built, the eight-task payload audit, and a recorded production trajectory.

\begin{table}[ht]
\centering
\caption{\textbf{The seven production primitives of VibeVideoAgent.} The first six call the decoupled perception worker on bounded excerpts; \texttt{probe\_media} runs deterministically on the local host and makes no model calls.}
\label{tab:primitives}
\small
\begin{tabularx}{\linewidth}{@{}l l l X@{}}
\toprule
\textbf{Role} & \textbf{Primitive} & \textbf{Verdict set} & \textbf{Primary in-production purpose} \\
\midrule
\multirow{4}{*}{\makecell[l]{I. Coarse-to-fine\\exploration}}
  & \texttt{describe\_media} & \texttt{observed}, \texttt{uncertain} & $1\,$fps global timeline index or still layout overview. \\
  & \texttt{locate\_event} & \texttt{observed}, \texttt{uncertain} & Query search plus automatic $8\,$fps refinement on $\le 2$ windows. \\
  & \texttt{inspect\_segment} & \texttt{observed}, \texttt{uncertain} & Focused local or cropped Q\&A ($4\,$fps default) plus $\le 6$ evidence frames. \\
  & \texttt{transcribe\_speech} & \texttt{observed}, \texttt{uncertain} & Audio-only speech transcription in the original language with timestamps. \\
\midrule
\makecell[l]{II. Sourced\\contract check}
  & \texttt{check\_requirement} & \texttt{pass}, \texttt{fail}, \texttt{uncertain} & Tests one sourced claim with enforced multi-stream evidence. \\
\midrule
\makecell[l]{III. Dual-clock\\comparison}
  & \texttt{compare\_media} & \makecell[l]{\texttt{resolved}, \texttt{persists},\\\texttt{changed}, \texttt{unchanged},\\\texttt{uncertain}} & Aligns two independent clocks for reference-to-draft matching or \texttt{finding\_id}-anchored revision verification. \\
\midrule
\makecell[l]{IV. Signal\\probing}
  & \texttt{probe\_media} & measured JSON & Local stream specifications, strict decode, LUFS loudness, $-50\,$dB silence, repeated-frame runs, and RGB loop-seam difference. \\
\bottomrule
\end{tabularx}
\end{table}

\paragraph{Additional primitive details.}
\texttt{locate\_event} returns uncertainty intervals around event boundaries. When \texttt{refine=True} and the scan is sparse ($f_{\text{eff}} < 8\,$fps), it extracts up to two candidate windows padded by $\pm 1/f_{\text{eff}}$ seconds and re-queries them at $8\,$fps, which resolves fast motion onsets and cut points without manual bookkeeping. \texttt{inspect\_segment} with \texttt{evidence\_frames=True} extracts up to six representative JPEG frames from the constant-frame-rate excerpt that the perception model saw and returns them to the coding agent for visual confirmation. \texttt{transcribe\_speech} drops all video tracks and uploads only lightweight Opus audio (\texttt{audio/ogg}); it returns verbatim utterances with start and end times, flags inaudible words, and answers locally without a model call if the container has no audio track. \texttt{check\_requirement} accepts a claim either directly (standalone mode) or by \texttt{requirement\_id} from the production plan. \texttt{compare\_media} requires evidence citations on both excerpts ($\{m_i : e_i \in \mathcal{V}\} = \{0, 1\}$), and in revision mode it requires the parent Finding's \texttt{required\_evidence} modalities on both excerpts. \texttt{probe\_media} also reports codec, resolution, frame rate, sample rate, channels, and duration. \texttt{plan\_production} returns deduplicated diagnostics in a single turn for requirements not linked to any beat, quotes that are not verbatim, and invalid \texttt{inputs/} paths.

\section{Complete MCP Tool Schemas}
\label{sec:app_schemas}

Every perceptual primitive in VibeVideoAgent accepts the unified \texttt{Media} object:
\begin{itemize}
    \item \texttt{path} (\texttt{str}, required): workspace-relative path to a supplied asset, still image, audio file, partial preview, or encoded video.
    \item \texttt{start}, \texttt{end} (\texttt{float}, default \texttt{0}, \texttt{None}): source-clock interval in seconds ($0 \le \texttt{start} < \texttt{end} \le \text{duration}$; omitted on stills).
    \item \texttt{timeline\_offset} (\texttt{float | None}): global film-timeline seconds corresponding to source time zero of a partial render.
    \item \texttt{modality} (\texttt{"auto" | "visual" | "audio" | "audiovisual"}, default \texttt{"auto"}): stream selection; \texttt{auto} inspects container tracks.
    \item \texttt{fps} (\texttt{float | None}, $0 < \texttt{fps} \le 16$): requested sampling rate, capped automatically by the configurable per-call frame budget ($B_{\text{frames}}=128$ in production workflow runs; $256$ schema default) and the native frame rate.
    \item \texttt{crop} (\texttt{[x, y, width, height] | None}): source-pixel region cropped \emph{before} downscaling.
    \item \texttt{source\_files} (\texttt{list[str]}, at most 32): workspace-relative editable source files hashed with SHA-256 at inspection time.
\end{itemize}
Table~\ref{tab:app_contracts} lists all production and management interfaces exposed by the VibeVideoAgent MCP server.

\begin{table}[ht]
\centering
\caption{\textbf{Complete MCP interface of VibeVideoAgent.} Management interfaces (\texttt{plan\_production}, \texttt{register\_artifact}, \texttt{submit\_delivery}, \texttt{status}) execute locally with zero model calls.}
\label{tab:app_contracts}
\small
\begin{tabularx}{\linewidth}{@{}l >{\raggedright\arraybackslash}p{4.6cm} >{\raggedright\arraybackslash}X@{}}
\toprule
\textbf{Tool} & \textbf{Key arguments} & \textbf{Returned payload and invariants} \\
\midrule
\texttt{plan\_production} & \texttt{goal}, \texttt{deliverables}, \texttt{requirements}, \texttt{beats}, \texttt{creative\_choices}, \texttt{first\_requirement}, \texttt{preview}, \texttt{risk} & Validates each verbatim \texttt{prompt\_quote} against \texttt{prompt.txt}, checks complete requirement-to-beat coverage (deduplicated diagnostics), and unlocks coding and rendering. \\
\addlinespace
\texttt{describe\_media} & \texttt{media}, \texttt{focus} & $1\,$fps sparse scene, text, and sound timeline or still layout; navigation only. \\
\addlinespace
\texttt{locate\_event} & \texttt{media}, \texttt{query}, \texttt{refine=True} & Candidate occurrences with boundary intervals plus up to two automatic $8\,$fps refinement Findings. \\
\addlinespace
\texttt{inspect\_segment} & \texttt{media}, \texttt{question}, \texttt{evidence\_frames=False} & $4\,$fps local diagnostic Finding; optionally attaches $\le 6$ CFR-sampled JPEG frames to the MCP response. \\
\addlinespace
\texttt{transcribe\_speech} & \texttt{media} & Extracts $64\,$kbps Opus audio only; returns verbatim segments with timestamps and a language tag (no API call if silent). \\
\addlinespace
\texttt{check\_requirement} & \texttt{requirement\_id}, \texttt{media}, \texttt{stage} $\in$ \{\texttt{preview}, \texttt{assembly}, \texttt{final}\} & Looks up the planned \texttt{observable}, \texttt{prompt\_quote}, and required \texttt{evidence} modalities; rejects \texttt{inputs/} paths; returns \texttt{pass}/\texttt{fail}/\texttt{uncertain}. \\
\addlinespace
\texttt{compare\_media} & \texttt{media}, (\texttt{reference} + \texttt{focus}) $\oplus$ \texttt{finding\_id} & Dual-clock comparison over $[E_0, E_1]$; inherits \texttt{required\_evidence}, tracks \texttt{changed\_declared\_sources}, and blocks same-SHA \texttt{resolved}. \\
\addlinespace
\texttt{probe\_media} & \texttt{path}, \texttt{measurements} $\subseteq$ \{\texttt{decode}, \texttt{motion}, \texttt{loudness}, \texttt{silence}\} & Local \texttt{ffprobe}/\texttt{ffmpeg} stream specifications, strict decode check, LUFS and true peak, $-50\,$dB silence spans, frozen runs, and RGB seam difference. \\
\addlinespace
\texttt{register\_artifact} & \texttt{video}, \texttt{source\_files}, \texttt{render\_command}, \texttt{role="output"} & Snapshots the immutable \texttt{0444} MP4 and source tree under \texttt{artifacts/vXXXX/} with a SHA-256 manifest (optional \texttt{repair\_id}, \texttt{runtime\_trace}). \\
\addlinespace
\texttt{submit\_delivery} & \texttt{artifact\_id=None}, \texttt{notes=""} & Verifies full-stream decode, delivery specifications, and preview-gate completion (or an explicit exception note); writes \texttt{verification.json}. \\
\addlinespace
\texttt{status} & (none) & Returns the recorded \texttt{production\_plan}, cumulative observation budget, registered artifacts, and \texttt{workflow\_checks} ledger (survives context compaction). \\
\bottomrule
\end{tabularx}
\end{table}

\section{Deterministic Finding Validation Rules}
\label{sec:app_validation}

Every structured JSON response from the perception worker (\texttt{media\_worker.py}) must conform to the schema $\{\texttt{summary}, \texttt{verdict}, \texttt{evidence}\}$ and pass five controller-enforced checks before entering \texttt{state.json}:
\begin{enumerate}
    \item \textbf{Excerpt index bounds.} Each evidence row's \texttt{media\_index} must reference a valid input excerpt ($0$ for single-media calls; $\{0,1\}$ for \texttt{compare\_media}).
    \item \textbf{Temporal and modality containment.} Evidence times must satisfy $0 \le \texttt{start} \le \texttt{end} \le \text{duration} + 10^{-3}$ on the referenced excerpt's clock, and \texttt{modality} $\in \{\texttt{visual}, \texttt{audio}\}$ must be compatible with the excerpt's extracted tracks.
    \item \textbf{Normalized box validity.} Non-empty boxes are permitted only on \texttt{visual} evidence and must satisfy $0 \le x_1 < x_2 \le 1$ and $0 \le y_1 < y_2 \le 1$.
    \item \textbf{Cross-excerpt and cross-modality completeness.} If any supplied excerpt receives no evidence rows, or any modality in \texttt{required\_evidence} lacks a cited evidence row on an excerpt, the \texttt{verdict} is deterministically downgraded to \texttt{uncertain}.
    \item \textbf{Comparison-mode and byte-identity guards.} Reference-mode \texttt{compare\_media} cannot emit revision verdicts (\texttt{resolved}/\texttt{persists}); revision-mode \texttt{compare\_media} emitting \texttt{resolved} when $\text{SHA256}(E_0^{\text{src}}) = \text{SHA256}(E_1^{\text{src}})$ is downgraded to \texttt{unchanged}.
\end{enumerate}

\section{Production Workflow Guidance and Cross-Stack Generality}
\label{sec:app_prompts}

\paragraph{General production workflow policy.}
In VibeVideoAgent runs, the system prompt is extended with short five-step guidance: (1)~read \texttt{prompt.txt}, \texttt{task.json}, and the input inventory, then call \texttt{plan\_production} to separate quoted requirements from creative choices and select the smallest render capable of testing the highest-risk decision; (2)~resolve unfamiliar input assets via \texttt{describe\_media}, \texttt{locate\_event}, \texttt{transcribe\_speech}, or \texttt{inspect\_segment}; (3)~render the minimal preview early and verify \texttt{first\_requirement} at \texttt{stage=preview} before scaling to the remaining beats; (4)~assemble the remaining beats, inspect cut boundaries and A--V synchronization at \texttt{stage=assembly}, and verify code repairs via \texttt{compare\_media(finding\_id, media)}; and (5)~run \texttt{probe\_media} and \texttt{stage=final} checks on the exact encoded delivery file before calling \texttt{register\_artifact} and \texttt{submit\_delivery}.

\paragraph{Zero-adapter operation across heterogeneous rendering pipelines.}
All primitives operate on standard images, audio streams, and video containers, and provenance can be bound to any source file, so no engine-specific adapters are needed: the seven MCP tool schemas and validation rules stay the same whichever rendering stack the coding agent chooses for a brief. (1)~In headless-browser and interactive 3D WebGPU/Canvas pipelines (Playwright, HTML/CSS, Three.js, React/Remotion), minimal previews capture single DOM or viewport stills or short frame-stepped clips to check layout legibility, 3D cutaway framing, and CSS/Canvas easing before full capture. (2)~In Python vector, scientific, and game-simulation pipelines (Cairo, Pillow, SVG, Manim), early previews check coordinate geometry, mathematical notation, and sprite state transitions, and \texttt{compare\_media(reference, focus)} aligns drafts with supplied reference visuals or motion specifications. (3)~In narrated and music-driven pipelines (\texttt{edge-tts}, SciPy synthesis, FFmpeg filtergraphs), \texttt{transcribe\_speech}, audiovisual \texttt{check\_requirement}, and \texttt{probe\_media(loudness, silence)} enforce speech-to-graphic timing and clean audio mixing without engine-specific plugins.

\section{Runtime Isolation and Post-Freeze Evaluation}
\label{sec:app_isolation}

Each task executes in an isolated Linux user, mount, and PID namespace with its own CLI runtime configuration directory (Claude Code for Claude models; Codex CLI for GPT-6 Astra) and locked MCP settings. The \texttt{perception\_gate.py} hook intercepts every \texttt{PreToolUse} event, logs its decision to \texttt{perception\_gate.jsonl}, and denies (i)~direct \texttt{Read} calls on binary image, audio, video, or PDF files; (ii)~sub-agent or alternative MCP invocations; (iii)~writes outside \texttt{workspace/} or inside the read-only \texttt{workspace/inputs/}; and (iv)~shell commands that attempt to invoke external model endpoints or inspect API credentials. Official benchmark rubrics and VLM or scene judges reside outside the task filesystem and are executed only after \texttt{run.json} reaches a terminal state and the delivered MP4 hash is frozen.

\section{Ablation Subset Construction}
\label{sec:app_subset}

More than $80\%$ of the benchmark tasks already reach $100\%$ L2 under Native Opus~5.5, so uniform random sampling squeezes the differences between variants against the ceiling, while selecting tasks by the gains of \textbf{+ VibeVideoAgent} would introduce post-hoc selection and regression-to-the-mean bias. We therefore build a deterministic, treatment-blind, capability-stratified subset of $36$ tasks ($15\%$ of VibeVideoBench) only from native-baseline diagnostics and public task metadata, before running any ablation. It has four strata:
\begin{itemize}
    \item \textbf{Rubric-bottleneck stratum ($12$ tasks):} multi-constraint briefs with Native L2 $< 98.0\%$, stressing upfront requirement decomposition and sourced contract verification.
    \item \textbf{Reference-grounded asset stratum ($8$ tasks):} tasks supplying external video, image, or audio assets in \texttt{inputs/} (sorted by lowest Native L3), directly exercising coarse-to-fine asset exploration and reference-mode \texttt{compare\_media}.
    \item \textbf{Audiovisual and temporal bottleneck stratum ($10$ tasks):} audio-mandated tasks with the lowest baseline L3 and \emph{AQ} scores, stressing modality-gated \texttt{check\_requirement}, \texttt{transcribe\_speech}, and \texttt{probe\_media}.
    \item \textbf{Median-baseline guardrail stratum ($6$ tasks):} standard tasks across major categories (\texttt{D}, \texttt{W}, \texttt{ZH}) with Native L2 $= 100\%$ and median Native L3 $\approx 4.00$, verifying that closed-loop inspection does not over-edit already compliant drafts.
\end{itemize}

\section{Eight-Task Payload Compression and Dual-Judge Audit}
\label{sec:app_pilot}

\begin{table}[ht]
\centering
\caption{\textbf{Eight-task diagnostic payload compression and dual-judge audit} (Claude Opus~5.5 + VibeVideoAgent in \texttt{CC 2.1.325} vs.\ the archived official Native Opus~5.5 rollout \texttt{v4\_claude\_opus\_sub} on VibeVideoBench). Tasks span five development cases (\texttt{W0448}, \texttt{W0100}, \texttt{D0032}, \texttt{D0054}, \texttt{D0060}) and three held-out hard cases selected before generation as the tasks with the lowest paired baseline pass rates (\texttt{W0154}, \texttt{R0009}, \texttt{PA02}). \emph{Calls (Find.) / Fr.} reports total perception-worker calls, Findings accepted after deterministic post-validation, and sampled frames; \emph{Perceived} and \emph{Returned} compare matched Gemini~3.8 Flash prompt tokens on accepted Findings ($38.1\text{k}$ mean across all worker attempts) with full serialized MCP Finding JSON tokens ($\approx 4\,$B/token; no evidence-frame JPEGs requested). L3 averages the 1--5 dimensions (\emph{AQ} omitted on brief-permitted silent deliveries \texttt{W0448} and \texttt{W0154}).}
\label{tab:pilot_audit}
\small
\setlength{\tabcolsep}{3.3pt}
\begin{tabularx}{\linewidth}{@{}X ccc cc cc@{}}
\toprule
& \multicolumn{3}{c}{\textbf{Perceptual context telemetry}} & \multicolumn{2}{c}{\textbf{Gemini 3.8 Flash judge}} & \multicolumn{2}{c}{\textbf{Gemini 4 Argon judge}} \\
\cmidrule(lr){2-4} \cmidrule(lr){5-6} \cmidrule(l){7-8}
\textbf{Task} & \textbf{Calls (Find.) / Fr.} & \textbf{Perceived} & \textbf{Returned} & \textbf{L2 (\%)} & \textbf{L3} & \textbf{L2 (\%)} & \textbf{L3} \\
\midrule
\texttt{W0448} & 13 (12) / 412   & 32.1k  & 5.8k ($5.5\times$)  & $76.0 \to \textbf{96.0}$ & $3.60 \to \textbf{4.75}$ & $95.8 \to 95.8$          & $3.60 \to \textbf{3.75}$ \\
\texttt{W0100} & 6 (5) / 180     & 15.2k  & 2.2k ($6.8\times$)  & $95.5 \to \textbf{100.0}$ & $3.60 \to \textbf{4.00}$ & $\textbf{89.6} \to 88.7$ & $3.20 \to \textbf{3.40}$ \\
\texttt{D0032} & 15 (11) / 243   & 16.7k  & 5.1k ($3.3\times$)  & $100.0 \to 100.0$        & $4.00 \to 4.00$          & $100.0 \to 100.0$        & $\textbf{5.00} \to 4.80$ \\
\texttt{D0054} & 17 (15) / 362   & 39.7k  & 8.4k ($4.7\times$)  & $100.0 \to 100.0$        & $4.00 \to 4.00$          & $100.0 \to 100.0$        & $4.00 \to 4.00$          \\
\texttt{D0060} & 14 (8) / 362    & 25.0k  & 4.5k ($5.6\times$)  & $100.0 \to 100.0$        & $5.00 \to 5.00$          & $100.0 \to 100.0$        & $3.80 \to \textbf{4.00}$ \\
\texttt{W0154}$^{\dagger}$ & 5 (4) / 100  & 9.1k  & 2.0k ($4.4\times$) & $62.7 \to \textbf{83.1}$ & $2.40 \to \textbf{3.75}$ & $13.8 \to \textbf{84.8}$ & $1.60 \to \textbf{3.50}$ \\
\texttt{R0009}$^{\dagger}$ & 14 (13) / 562 & 42.8k & 7.0k ($6.1\times$) & $92.7 \to 92.7$        & $3.60 \to \textbf{4.80}$ & $91.9 \to \textbf{92.1}$ & $3.60 \to \textbf{3.80}$ \\
\texttt{PA02}$^{\dagger}$  & 13 (13) / 1106 & 105.6k & 8.8k ($12.0\times$) & $94.5 \to \textbf{96.4}$ & $4.00 \to \textbf{5.00}$ & $89.3 \to \textbf{94.5}$ & $3.80 \to 3.80$       \\
\midrule
\textbf{Mean (8)} & \textbf{12.1 (10.1) / 416} & \textbf{35.8k} & \textbf{5.5k ($6.5\times$)} & $90.2 \to \textbf{96.0}$ & $3.77 \to \textbf{4.41}$ & $85.0 \to \textbf{94.5}$ & $3.58 \to \textbf{3.88}$ \\
\bottomrule
\end{tabularx}
\par\smallskip
\begin{minipage}{\linewidth}\footnotesize
$^{\dagger}$Held-out hard tasks selected before generation; L2/L3 show Native $\to$ Ours; the \texttt{stage=preview} gate is checked on $8/8$ ($100\%$) tasks.
\end{minipage}
\end{table}

Table~\ref{tab:pilot_audit} reports exact single-turn MCP response sizes on the eight-task Opus~5.5 diagnostic suite. The agent passes the upfront \texttt{plan\_production} and \texttt{stage=preview} gates on all eight tasks and requests no \texttt{evidence\_frames} attachments, so the coding context stays purely textual. Each task makes $12.1$ worker calls on average, of which $10.1$ yield accepted Findings; about $2.0$ calls per task are rejected by the deterministic Finding validator because the model's evidence falls outside the excerpt bounds. The short-lived worker calls take in $35{,}767$ multimodal prompt tokens per task (up to $105{,}606$ tokens over $1{,}106$ frames and $884\,$s of narration on the long-form paper explainer \texttt{PA02}) and return only $5{,}465$ serialized Finding tokens to the coding session. This is a $6.5\times$ reduction for a single pass ($12.0\times$ on \texttt{PA02}), before any replay of multi-turn history. On these harder diagnostic cases, VibeVideoAgent raises L2 from $90.2\%$ to $96.0\%$ (L3 $3.77 \to 4.41$) under Gemini~3.8 Flash and from $85.0\%$ to $94.5\%$ (L3 $3.58 \to 3.88$) under Gemini~4 Argon, including a $71.1\,$pp gain ($13.8\% \to 84.8\%$) on the held-out hard task \texttt{W0154}.

\section{Recorded Production Trajectory}
\label{sec:app_case}

\begin{figure}[ht]
  \centering
  \includegraphics[width=\linewidth]{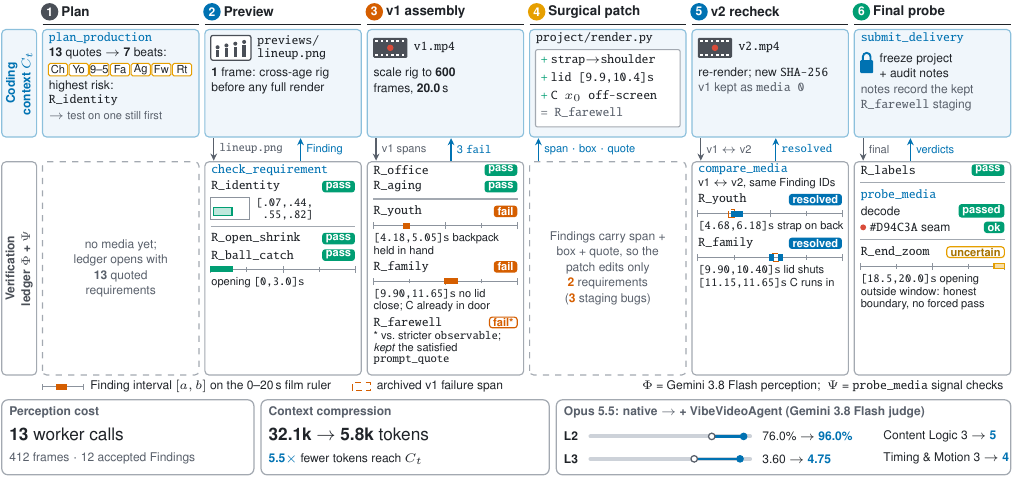}
  \caption{\textbf{Recorded production trajectory on \texttt{W0448}} (a $20.0\,$s, $1920{\times}1080$, $30\,$fps six-stage life-story loop). Every timestamp, normalized box, and verdict is copied from \texttt{state.json}; Finding spans are drawn on a 0--20\,s film ruler. The agent validates the cross-age character design on one preview still and catches three silent staging bugs in \texttt{R\_youth} and \texttt{R\_family} in \texttt{v1.mp4}. It also separates an over-specific \texttt{observable} it wrote for \texttt{R\_farewell} (marked fail*) from the satisfied \texttt{prompt\_quote}. It then patches \texttt{render.py}, confirms both repairs with dual-clock \texttt{compare\_media}, and finishes with decode and seam probes, leaving \texttt{R\_end\_zoom} as \texttt{uncertain}. The bottom row reports perception cost, context compression, and the Flash-judged outcome against the native Opus~5.5 baseline.}
  \label{fig:trajectory}
\end{figure}

Figure~\ref{fig:trajectory} traces one complete recorded run of VibeVideoAgent (Claude Opus~5.5) on VibeVideoBench task \texttt{W0448}. The task asks for a $20.0\,$s ($1920{\times}1080$, $30\,$fps) wordless life story in six stages (\emph{Childhood}, \emph{Youth}, \emph{9--5}, \emph{Family}, \emph{Aging}, \emph{Farewell}), anchored by a recurring red button (\texttt{\#D94C3A}) that opens and closes the film as a seamless loop. Before any rendering code is written, \texttt{plan\_production} extracts 13 verbatim-quoted requirements across 7 beats and marks multi-age character identity (\texttt{R\_identity}) as the highest-risk decision. The agent tests it first on a single rendered lineup still (\texttt{previews/lineup.png}) with \texttt{check\_requirement(stage=preview)}, which returns \texttt{pass} (box $[0.07, 0.44, 0.55, 0.82]$). The agent then scales the verified rig to the first 600-frame film \texttt{v1.mp4}, which passes the opening, office, and aging checks, but assembly-stage checks return three \texttt{fail} verdicts. In \emph{Youth} ($[4.18, 5.05]\,$s, box $[0.69, 0.48, 0.81, 0.82]$), the protagonist catches the backpack and holds it down in hand instead of slinging it onto the shoulder. In \emph{Family} ($[9.90, 11.65]\,$s), no laptop lid closes before the character stands ($[9.90, 10.65]\,$s), and child~C is already waiting inside the doorway instead of running in from off-screen ($[10.90, 11.65]\,$s). In \emph{Farewell} ($[17.38, 19.00]\,$s), the coat hangs on a stand \emph{beside} the empty chair. This fails the stricter \texttt{observable} that the agent wrote itself (\emph{``draped on it''}) but satisfies the verbatim \texttt{prompt\_quote} (\emph{``Beside an empty chair, place P's coat''}).

\begin{figure}[ht]
  \centering
  \includegraphics[width=\linewidth]{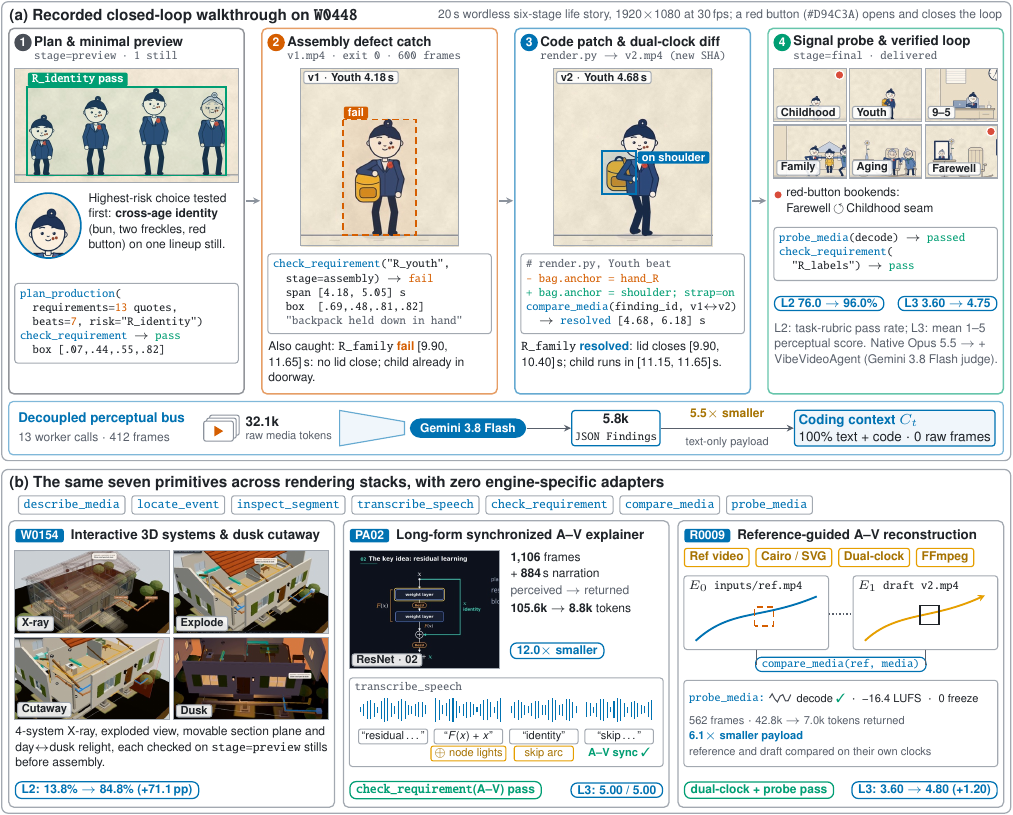}
  \caption{\textbf{End-to-end authoring with VibeVideoAgent.} \textbf{(a)}~The recorded \texttt{W0448} run with real frames. (1)~\texttt{plan\_production} decomposes the brief, and a single \texttt{stage=preview} lineup still verifies cross-age character identity before full rendering. (2)~\texttt{stage=assembly} checks catch silent staging bugs in \texttt{v1.mp4} even though the render exited cleanly: the backpack is held in hand, and in the \emph{Family} beat no lid closes while the child already stands in the doorway. (3)~\texttt{compare\_media(finding\_id, media)} confirms a targeted \texttt{render.py} patch across shifted clocks. (4)~Deterministic \texttt{probe\_media} checks finalize the seamless red-button loop. Raw frames stay inside short-lived Gemini~3.8 Flash calls, whose Findings are $5.5\times$ smaller than the media they perceive. \textbf{(b)}~Without engine-specific adapters, the same seven primitives serve an interactive 3D WebGPU system (\texttt{W0154}, $+71.1\,$pp L2 under the Argon judge), a long-form paper explainer with synchronized narration (\texttt{PA02}, $12.0\times$ smaller payload), and reference-guided reconstruction of motion and audio (\texttt{R0009}, Flash-judged L3 $3.60 \to 4.80$).}
  \label{fig:hero}
\end{figure}

Because \texttt{check\_requirement} keeps both strings, the agent keeps the Farewell staging that follows the prompt (and notes this in \texttt{submit\_delivery.notes}). It patches \texttt{project/render.py} for the three real bugs in \emph{Youth} and \emph{Family}, renders \texttt{v2.mp4}, and calls \texttt{compare\_media} with the earlier Finding IDs. Both dual-clock rechecks return \texttt{resolved} against the archived \texttt{v1} excerpts ($[4.68, 6.18]\,$s for \texttt{R\_youth}; $[9.90, 10.40]\,$s and $[11.15, 11.65]\,$s for \texttt{R\_family}). The agent then checks \texttt{R\_end\_zoom} at \texttt{stage=final} on the closing excerpt $[18.50, 20.00]\,$s alone, after an earlier call on $[18.60, 20.00]\,$s was rejected by the deterministic post-validator. \texttt{check\_requirement} confirms the zoom into red ($[19.13, 19.88]\,$s) but returns \texttt{uncertain} instead of guessing a pass, because the opening zoom-out lies outside that window. \texttt{probe\_media(decode)} passes, and a local RGB seam check confirms the loop. The run uses $13$ perception-worker calls ($412$ frames, $12$ accepted Findings), with $32{,}079$ prompt tokens for accepted Findings ($32{,}320$ in total) against $5{,}811$ serialized Finding tokens and $4{,}561$ model output tokens. Under the Gemini~3.8 Flash judge, it raises the task-rubric pass rate from $76.0\%$ ($19/25$) for the native Opus~5.5 baseline to $96.0\%$ ($24/25$), with Content Logic rising from $3$ to $5$ and Timing \& Motion from $3$ to $4$. Figure~\ref{fig:hero} shows the same run with real frames, together with three other tasks that use the same seven primitives on different rendering stacks.

\end{document}